\documentclass[conference,letterpaper]{IEEEtran}

\usepackage[
  letterpaper,
  top=0.74in,
  bottom=0.74in,
  left=0.75in,
  right=0.75in
]{geometry}

\usepackage{amsmath,amssymb,amsfonts}
\usepackage{algorithmic}
\usepackage[ruled,vlined,linesnumbered]{algorithm2e}
\usepackage{graphicx}
\usepackage{textcomp}
\usepackage{xcolor}
\usepackage{booktabs}
\usepackage{multirow}
\usepackage{cite}
\usepackage{url}
\usepackage{balance}
\usepackage[hidelinks]{hyperref}
\usepackage{enumitem}

\IEEEoverridecommandlockouts

\begin{document}

\title{Autonomous Model Lifecycle Management for Digital Twin-Based Manufacturing Control}

\author{Zhengyang ``Cissy'' Gu$^{*}$, Thomas Cook, Fredaljohn Rohrbaugh, Joseph E. Hernandez, and Chris Couch%
\thanks{All authors are with Liveline Technologies, Livonia, MI, USA.}%
\thanks{This project was funded by Liveline Technologies. Liveline Technologies is dedicated to improving manufacturing performance by harnessing the power of Artificial Intelligence to automate complex processes and predict future problems.}%
\thanks{This is the author's version of this work. It has been accepted for publication in the 2026 IEEE International Conference on Systems, Man, and Cybernetics (SMC 2026). The final published version will be made available on IEEE Xplore.}%
\thanks{\copyright~2026 IEEE. Personal use of this material is permitted. Permission from IEEE must be obtained for all other uses, in any current or future media, including reprinting/republishing this material for advertising or promotional purposes, creating new collective works, for resale or redistribution to servers or lists, or reuse of any copyrighted component of this work in other works.}%
}

\maketitle

\begin{abstract}
Manufacturing AI systems must autonomously adapt to continuous distributional shift from raw-material variability, ambient changes, and equipment aging, under strict safeguard and operator-trust requirements where model failures risk physical damage. This paper presents a closed-loop Cyber-Physical System (CPS) for autonomous model lifecycle management in automotive manufacturing, deployed since 2023. The system manages product-specialized model pairs: a sequence-to-sequence physics model (LPP) serving as a digital twin, and a deep Reinforcement Learning (RL) control policy (LCP) trained against it. Per retraining cycle, multiple model variants spanning architecture families and RL algorithms compete; only the best-scoring candidate advances. A \emph{Conductor} orchestrator autonomously manages plant-wide model inventories with dependency-aware retraining and Proportional-Integral-Derivative (PID) fallback. Reflecting the principle of \emph{Human-Centric Intelligence}, the LCP composite score embeds an \emph{operator-trust gate} penalizing policies deviating from established practice; without it, 23\% of policies are rejected by operators despite passing accuracy thresholds. Across multiple facilities, LCP-controlled processes achieve process stability improvements of 28--45\% over uncontrolled baselines with zero safety incidents.
\end{abstract}


\section{Introduction}
\label{sec:intro}

Each product at a manufacturing facility requires a dedicated model pair: a physics model predicting process outputs and a control policy recommending setpoints. These models degrade due to material variability, tooling wear, and environmental shifts (Section~\ref{sec:background}), and when a physics model is retrained, the dependent control policy must follow. A plant with 20 products across 10 lines must manage 200+ such pairs, making manual lifecycle management infeasible~\cite{lee2018industrial}. The lifecycle must be \emph{autonomous}: sensing degradation, qualifying data, selecting the best candidate from competing variants, enforcing dependency ordering, gating deployment, and falling back to safe classical control, all without human intervention.

This paper addresses \emph{autonomous model lifecycle management} for safety-critical manufacturing AI. Unlike web-scale Machine Learning Operations (MLOps) where updates are low-risk, manufacturing model failures risk physical damage and safety incidents~\cite{paleyes2022challenges}. The system guarantees: (1)~no model reaches production without passing automated backtesting; (2)~a deterministic PID fallback operates at all times; and (3)~control policies respect operator trust~\cite{hancock2021evolving}.

Our contributions are: (1)~a closed-loop CPS architecture with a seven-stage MLOps cycle (Fig.~\ref{fig:lifecycle}); (2)~\emph{competitive model selection} where multiple architecture families and RL algorithms compete per retraining cycle; (3)~composite evaluation using $C_{pk}$ (process stability index), Spearman correlation, and an explicit operator-trust term; (4)~a five-layer human-centric safeguard architecture; and (5)~quantitative production evaluation demonstrating $C_{pk}$ improvements of 28--45\%.

\section{Background and Related Work}
\label{sec:background}

\subsection{Manufacturing-Specific Challenges}

Manufacturing imposes challenges distinct from web-scale ML. \textbf{Material variability:} polymer viscosity, filler content, and moisture vary across suppliers; material lot transitions cause discrete, unpredictable distribution shifts~\cite{paleyes2022challenges}. \textbf{Tooling wear:} dies, heaters, and actuators degrade continuously, then change abruptly at maintenance. \textbf{Ambient environment:} factory temperature, humidity, and pressure cause seasonal drift and acute transients. \textbf{Signal noise:} Electromagnetic Interference (EMI), Programmable Logic Controller (PLC) dropouts, and sensor miscalibration corrupt data. These collectively motivate \emph{continuous autonomous retraining} and \emph{layered human-centric safeguards}.

\subsection{Related Work}

Digital twin technology is foundational to smart manufacturing~\cite{tao2019digital}, with extensions to cognitive twins~\cite{zheng2021emerging} and real-time synchronization~\cite{qi2021enabling}, though most work targets simulation rather than closed-loop autonomous retraining. Deep learning has been widely applied to manufacturing tasks including quality prediction and fault detection~\cite{wang2018deep}, yet few systems address the full model lifecycle. Deep RL for industrial control~\cite{spielberg2019toward} requires high-fidelity simulators, safety fallbacks, and adaptation to non-stationary dynamics; safe RL surveys~\cite{garcia2015safe} highlight the gap between RL theory and deployment in safety-critical domains. Existing MLOps frameworks~\cite{kreuzberger2023mlops, zaharia2018mlflow} lack manufacturing-specific capabilities such as product-specialized inventories, dependency-aware retraining, $C_{pk}$-based gating, and deterministic fallback~\cite{sculley2015hidden, polyzotis2018lifecycle, breck2019data}. Cyber-physical production systems~\cite{monostori2014cyber} provide the architectural foundation for integrating ML into physical processes, but do not address autonomous model lifecycle management. Trust in autonomous systems is a recognized IEEE Systems, Man, and Cybernetics (SMC) concern~\cite{hancock2021evolving, meng2022wavelet}; our system operationalizes trust as a measurable scoring constraint.

\section{System Design}
\label{sec:system}

\subsection{Closed-Loop MLOps Lifecycle}

\begin{figure}[t]
\centering
\includegraphics[width=0.85\columnwidth]{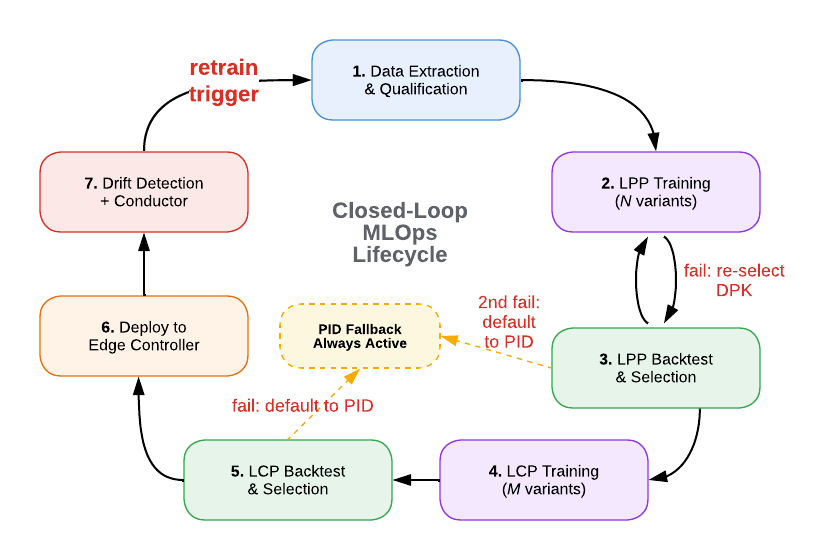}
\caption{Seven-stage closed-loop MLOps lifecycle. LPP backtest (Stage~3) gates LCP training; on failure, data packages are re-selected (max 2 attempts). If both fail, the system defaults to PID. LCP backtest (Stage~5) gates deployment. PID fallback remains active regardless of ML state.}
\label{fig:lifecycle}
\end{figure}

The system implements a seven-stage closed-loop MLOps lifecycle (Fig.~\ref{fig:lifecycle}) purpose-built for manufacturing. Each product is managed through a pair of coupled models: a \emph{Liveline Physics Package} (LPP), a sequence-to-sequence digital twin predicting process outputs, and a \emph{Liveline Controls Package} (LCP), a deep RL control policy recommending optimal setpoints using the LPP as its simulator (detailed in Sections~\ref{sec:modelpairs} and~\ref{sec:models}). Unlike linear ML pipelines that terminate at deployment, our architecture forms a continuous cycle where deployed-model performance feeds back to trigger the next iteration.

The seven stages are:
\textbf{(1)~Data Extraction \& Qualification:} sensor data is extracted via OPC-UA and transformed into proprietary Data Packages (DPK) with signal alignment, outlier detection, and train/test splitting; packages undergo data qualification (Section~\ref{sec:models}).
\textbf{(2)~LPP Training:} $N$ physics model variants spanning architecture families (LSTM, Transformer, Mamba) are trained on GPU compute.
\textbf{(3)~LPP Backtest \& Selection:} all $N$ variants are evaluated on held-out production runs; the best candidate is selected. On failure, data packages are re-selected and training retried; a second failure aborts the cycle and defaults to PID.
\textbf{(4)~LCP Training:} only after LPP passes, $M$ policy variants spanning RL algorithms (DDPG, TD3, SAC, PPO) are trained against the validated LPP.
\textbf{(5)~LCP Backtest \& Selection:} all $M$ variants are evaluated; on failure, the system defaults to PID.
\textbf{(6)~Deployment:} promoted models deploy to edge controllers; PID fallback remains active.
\textbf{(7)~Drift Detection \& Conductor:} the Conductor evaluates plant-wide inventories weekly and triggers the next cycle on drift or new data.

\subsection{Product-Specialized Model Pairs}
\label{sec:modelpairs}

Each product $p \in \mathcal{P}$ has a dedicated model pair $(\mathcal{M}_p^{\text{LPP}}, \mathcal{M}_p^{\text{LCP}})$: the LPP learns forward dynamics and the LCP learns a control policy using the LPP as its simulator. For a facility with lines $\mathcal{L}$ and products $\mathcal{P}$, the Conductor manages
\begin{equation}
\mathcal{I} = \{(\mathcal{M}_{p,l}^{\text{LPP}}, \mathcal{M}_{p,l}^{\text{LCP}}) \mid p \in \mathcal{P},\; l \in \mathcal{L}_p \}.
\label{eq:inventory}
\end{equation}

\subsection{Digital Twin and Control Models}
\label{sec:models}

\textbf{LPP (Seq2Seq Physics Model).} The LPP maps $T_{\text{in}}{=}120$ steps of $d_{\text{in}}$ input signals (e.g., barrel temperatures, screw speed) to $T_{\text{out}}{=}30$ predicted steps of $d_{\text{out}}$ outputs (e.g., melt temperature, part dimensions):
$\mathbf{X} \in \mathbb{R}^{T_{\text{in}} \times d_{\text{in}}} \to \hat{\mathbf{Y}} \in \mathbb{R}^{T_{\text{out}} \times d_{\text{out}}}$.
The architecture is not fixed: each cycle trains variants from LSTM encoder-decoder, Transformer, and Mamba backbones, with and without teacher forcing. The data pipeline applies filtering, outlier detection, dual-pass smoothing, and min-max normalization.

The LSTM variant uses an encoder that compresses the input into a hidden state $\mathbf{h}_{T_{\text{in}}}^{\text{enc}} = \text{LSTM}_{\text{enc}}(\mathbf{x}_{1:T_{\text{in}}})$, and a decoder that generates predictions autoregressively from this state:
\begin{equation}
\hat{\mathbf{y}}_t = f_{\text{dense}}\!\left(\text{LSTM}_{\text{dec}}(\hat{\mathbf{y}}_{t-1}, \mathbf{h}_{t-1}^{\text{dec}})\right),\quad \mathbf{h}_0^{\text{dec}} = \mathbf{h}_{T_{\text{in}}}^{\text{enc}}
\end{equation}
where $f_{\text{dense}}$ projects to the $d_{\text{out}}$-dimensional output space. Transformer and Mamba variants use analogous encoder-decoder structures.

\textbf{LCP (Deep RL Control Policy).} The LCP learns $\pi_\theta: \mathcal{S} \to \mathcal{A}$ mapping states (sensor readings) to actions (setpoint adjustments) via deep RL~\cite{spielberg2019toward} against the LPP. Multiple algorithms (DDPG, TD3, SAC, PPO) each produce a competing variant. The reward combines setpoint tracking with constraint satisfaction:
\begin{equation}
r_t = \sum_{j=1}^{d_{\text{out}}} \left[ \underbrace{-\alpha_j \left(\hat{y}_{t,j} - \tau_j\right)^2}_{\text{setpoint penalty}} - \underbrace{\beta_j \cdot \mathbf{1}\!\left(\hat{y}_{t,j} \notin [L_j, U_j]\right)}_{\text{constraint penalty}} \right]
\label{eq:reward}
\end{equation}
where $\hat{y}_{t,j}$ is the predicted output, $\tau_j$ the setpoint, and $[L_j, U_j]$ the spec limits. The constraint penalty is necessary because the smooth quadratic term alone does not enforce a hard boundary. Both $\alpha_j$, $\beta_j$ are configured per product by domain experts.

\textbf{Data Qualification.} The overlap ratio between the spec interval $[L_j, U_j]$ and observed interquartile range $[Q_{25,j}, Q_{75,j}]$ is:
\begin{equation}
\eta_j = \frac{\max\!\left(0,\; \min(U_j, Q_{75,j}) - \max(L_j, Q_{25,j})\right)}{U_j - L_j}.
\label{eq:overlap}
\end{equation}
Sufficient overlap ensures data contains both in-spec samples (positive rewards) and near-boundary samples (negative penalties) for boundary-aware RL training. Packages are accepted only if $\eta_{\text{min}} \leq \eta_j \leq \eta_{\text{max}}$ for all outputs, retain ${\geq}20\%$ of samples after outlier removal, and contain ${\geq}1{,}000$ samples.

\subsection{Competitive Model Selection}
\label{sec:competition}

\begin{figure}[t]
\centering
\includegraphics[width=0.85\columnwidth, height=5.5cm, keepaspectratio]{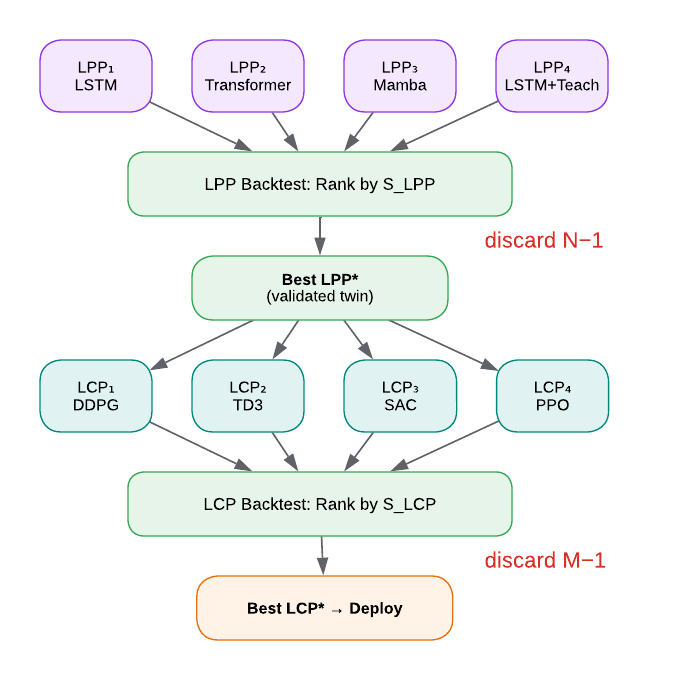}
\caption{Competitive model selection. $N$ LPP variants and $M$ LCP variants are trained per cycle. Only the best-scoring candidate advances; others are discarded.}
\label{fig:competition}
\end{figure}

\emph{No single architecture or algorithm is universally optimal} across products and conditions. Per cycle, $N$ LPP variants and $M$ LCP variants are independently backtested; only the best-scoring candidate at each stage advances (Fig.~\ref{fig:competition}). This tournament-style selection adapts automatically to non-stationary manufacturing dynamics without manual algorithm selection. For example, the winning architecture for a given product may shift between retraining cycles as ambient conditions change seasonally or after equipment maintenance alters process dynamics.

\subsection{Conductor Orchestrator}

The Conductor (Algorithm~\ref{alg:conductor}) runs weekly per facility. Retraining triggers on: (i)~performance degradation beyond facility-specific thresholds; (ii)~equipment changes such as tooling replacement or maintenance; or (iii)~accumulation of sufficient new qualified data since the last training cycle. Distribution shift is monitored using Population Stability Index (PSI) and Kolmogorov-Smirnov (KS) tests on incoming sensor data relative to the training distribution. The dependency chain is strict: LCP retraining requires the LPP to first pass backtesting. Partial retraining is supported: if only LCP has degraded while the LPP remains valid, LCP retrains against the validated incumbent LPP without re-triggering LPP training.

\begin{algorithm}[t]
\small
\DontPrintSemicolon
\SetAlgoLined
\SetKwInOut{Input}{Input}
\Input{Facility $f$, inventory $\mathcal{I}_f$}
\ForEach{line $l \in \mathcal{L}_f$}{
    \ForEach{product $p \in \mathcal{P}_l$}{
        $(m_{\text{lpp}}, m_{\text{lcp}}) \gets$ \textsc{GetModels}$(l, p)$\;
        $(\boldsymbol{\mu}, e, \mathcal{D}) \gets$ \textsc{GetStatus}$(m_{\text{lpp}}, m_{\text{lcp}}, l, p)$\;
        \If{\textsc{NeedsRetrain}$(m_{\text{lpp}}, \boldsymbol{\mu}, e, \mathcal{D})$}{
            Train $N$ LPP variants $\to$ backtest best\;
            \lIf{pass}{Train $M$ LCP variants $\to$ backtest best}
        }
        \ElseIf{\textsc{NeedsRetrain}$(m_{\text{lcp}}, \boldsymbol{\mu}, e, \mathcal{D})$}{
            Train $M$ LCP variants (current LPP) $\to$ backtest best\;
        }
    }
}
\caption{Conductor with Competitive Selection}
\label{alg:conductor}
\end{algorithm}

\section{Evaluation}
\label{sec:eval}

\subsection{Scoring Framework}

\textbf{LPP metrics} (per held-out run, rolling window~60, min-periods~30): Spearman $\rho_j$ for monotonic relationships; Root Median Squared Error $\text{RMdSE}_j = \sqrt{\text{median}((\bar{\hat{y}}_{i,j} - \bar{y}_{i,j})^2)}$; and Mean Directional Accuracy $\text{MDA}_j$ for trend correctness. Median-based metrics are preferred over mean-based alternatives for robustness to outliers common in manufacturing sensor data.

\textbf{LPP Composite Score} (lower is better):
\begin{equation}
S_{\text{LPP}} = 0.5\,(1 - \bar{\rho}) + 0.3\,\overline{\text{RMdSE}} + 0.2\,(1 - \overline{\text{MDA}}).
\label{eq:slpp}
\end{equation}

\textbf{LCP metrics:} Process capability $C_{pk,j} = \min\!\left(\frac{U_j - \hat{\mu}_j}{3\hat{\sigma}_j},\; \frac{\hat{\mu}_j - L_j}{3\hat{\sigma}_j}\right)$; setpoint deviation $\Delta_{\text{sp},i} = \mathbb{E}[\bar{a}_{t,i}^{\text{suggested}} - \bar{a}_{t,i}^{\text{observed}}]$; cumulative reward $\bar{R}$.

\textbf{LCP Composite Score} (higher is better):
\begin{equation}
S_{\text{LCP}} = 0.3\,\bar{R} + 0.2\,\overline{C}_{pk} - 0.5\,|\overline{\Delta}_{\text{sp}}|.
\label{eq:slcp}
\end{equation}

The 0.5 weight on $|\overline{\Delta}_{\text{sp}}|$ is the \emph{operator-trust term}. Weights were calibrated via grid search on one facility using operator-acceptance labels, then held fixed.

\subsection{Operational Results}

The system has been in production across multiple automotive facilities since 2023, managing over 200 active model pairs across 10 production lines. LPP and LCP each require 1--2~hours of training per product on GPU instances. The pipeline failure rate is ${\sim}$2\%, with all failed messages automatically re-driven from dead-letter queues. Data qualification rejects 8--12\% of packages. The Conductor averages 3.2 retraining cycles per product per quarter, of which 32\% are triggered by performance degradation and 68\% by equipment or process changes; 78\% of retrained models pass backtesting. No safety incidents have been attributed to ML-controlled operation. Both PID and LCP significantly outperform uncontrolled operation; LCP further improves upon tuned PID in $C_{pk}$ when a validated model is available, while PID serves as the reliable baseline during model transitions. In a controlled 6-month evaluation at one facility, disabling the operator-trust gate ($|\overline{\Delta}_{\text{sp}}|$ penalty) resulted in 23\% of deployed policies being overridden by operators despite meeting all accuracy and $C_{pk}$ thresholds, confirming that operator acceptance requires explicit trust-aware scoring.

\subsection{Baseline Comparison}

Table~\ref{tab:performance} compares three control regimes on identical production runs: \emph{Tuned PID} (incumbent classical controller), \emph{Uncontrolled} (LPP forward simulation with observed setpoints), and \emph{LCP} (deep RL policy selected via competitive evaluation). The LCP achieves the highest $C_{pk}$ for the majority of the measured outputs, with the largest gains on outputs exhibiting high variance under uncontrolled operation.

\begin{table}[t]
\centering
\caption{LPP Accuracy and LCP Control Performance. {\footnotesize\textit{Averaged across 3 facilities; $\uparrow$ higher better, $\downarrow$ lower better.}}}
\label{tab:performance}
\footnotesize
\begin{tabular}{@{}lccc@{}}
\toprule
\textbf{LPP Metrics} & \multicolumn{3}{c}{\textbf{Mean $\pm$ Std}} \\
\midrule
Spearman $\rho$ $\uparrow$        & \multicolumn{3}{c}{0.72 $\pm$ 0.13} \\
RMdSE $\downarrow$                & \multicolumn{3}{c}{0.05 $\pm$ 0.02} \\
MDA $\uparrow$                    & \multicolumn{3}{c}{0.78 $\pm$ 0.15} \\
$S_{\text{LPP}}$ $\downarrow$     & \multicolumn{3}{c}{0.20 $\pm$ 0.10} \\
\midrule
\textbf{Control Metric}    & \textbf{Uncntrl.} & \textbf{PID} & \textbf{LCP (ours)} \\
\midrule
Avg $C_{pk}$ $\uparrow$           & 2.63 & 3.05 & \textbf{3.37-3.81} \\
$\Delta C_{pk}$ vs.\ uncntrl.     & N/A  & 16\% & \textbf{+28--45\%} \\
$\Delta C_{pk}$ vs.\ PID          & N/A  & N/A  & \textbf{+10--25\%} \\
Constraint viol.                   & 7\% & 0\% & \textbf{0\%} \\
\bottomrule
\end{tabular}
\end{table}

\section{Discussion}
\label{sec:discussion}

\subsection{Five-Layer Human-Centric Safeguards}

In order of increasing criticality:
\textbf{L5~Pipeline Fault Isolation:} every queue is paired with a Dead-Letter Queue; ${\sim}2\%$ of messages reach DLQs and are automatically re-driven.
\textbf{L4~Data Qualification:} overlap analysis (Eq.~\ref{eq:overlap}) rejects 8--12\% of packages not representing the target regime.
\textbf{L3~Operator Trust Gate:} the LCP composite (Eq.~\ref{eq:slcp}) assigns 50\% weight to $|\overline{\Delta}_{\text{sp}}|$, operationalizing trust as a measurable constraint~\cite{hancock2021evolving}. Even a theoretically superior policy will fail in practice if operators do not trust it and override its recommendations.
\textbf{L2~Deployment Gating:} a retrained model must outperform the incumbent on aggregate metrics \emph{and} meet absolute thresholds before promotion.
\textbf{L1~PID Fallback:} a classical PID controller operates continuously in parallel; if the LCP is unavailable, produces out-of-range setpoints, or is overridden by an operator, PID assumes control deterministically with no interruption to production.

\subsection{Threats to Validity}

\emph{T1:} All telemetry comes from one company's polymer deployments; patterns may not transfer to other materials or industries.
\emph{T2:} Figures derive from internal dashboards, not third-party audit.
\emph{T3:} $C_{pk}$ is computed against fixed limits; tighter limits would reduce absolute values, though relative improvements should be preserved.
\emph{T4:} The 23\% operator-rejection rate was measured at a single facility over 6 months ($n{=}$15 policies evaluated); the direction should generalize but magnitude may vary across facilities and operator populations.

\subsection{Lessons Learned}

Per-facility threshold tuning is critical: retraining criteria at one plant may be too sensitive or lenient at another due to differences in equipment age, material suppliers, and ambient conditions. We maintain per-facility configurations with a calibration workflow using initial production data. The PID fallback has been activated during model transition windows and during an early deployment where LCP setpoints exceeded operator comfort ranges; in all cases, operations continued uninterrupted. We also observed that data qualification thresholds require periodic review as product specifications evolve, to avoid silently rejecting valid training data.

\section{Conclusion}
\label{sec:conclusion}

We presented a Cyber-Physical System approach to autonomous model lifecycle management for digital twin-based manufacturing control. The seven-stage closed-loop MLOps lifecycle with competitive model selection enables $N$ architecture variants and $M$ RL algorithm variants to compete per retraining cycle, with an autonomous Conductor managing dependency-aware retraining across plant-wide inventories. The operator-trust gate operationalizes trust as a measurable scoring constraint; without it, 23\% of accuracy-passing policies are rejected by operators. Production deployments since 2023 demonstrate $C_{pk}$ improvements of 28--45\% over uncontrolled baselines and zero ML-attributed safety incidents under five-layer safeguards. Future work includes Bayesian change-point detection for more responsive retraining triggers, federated learning across facilities to share model insights while preserving data locality, and automated root-cause analysis correlating model degradation with upstream process changes such as material lot transitions or maintenance events.

\balance

\end{document}